\pdfoutput=1

\documentclass[11pt]{article}

\usepackage{acl}

\usepackage{times}
\usepackage{latexsym}

\usepackage[T1]{fontenc}

\usepackage[utf8]{inputenc}

\usepackage{microtype}

\usepackage{array}
\usepackage{multirow}
\usepackage{amssymb}
\usepackage{amsmath}
\usepackage{mathrsfs}
\usepackage{graphicx}
\usepackage{booktabs}
\usepackage{url}
\usepackage{subfigure}
\usepackage{pifont}

\usepackage{enumitem}
\setenumerate[1]{itemsep=3pt,partopsep=0pt,parsep=\parskip,topsep=3pt}
\setitemize[1]{itemsep=3pt,partopsep=0pt,parsep=\parskip,topsep=3pt}
\setdescription{itemsep=3pt,partopsep=0pt,parsep=\parskip,topsep=3pt}

\title{Pretrained Language Encoders are Natural Tagging Frameworks for Aspect Sentiment Triplet Extraction}

\author{Yanjie Gou\textsuperscript{\rm 1}, 
Yinjie Lei\textsuperscript{\rm 1}, Lingqiao Liu\textsuperscript{\rm 2}, Yong Dai\textsuperscript{\rm 3}, Chunxu Shen\textsuperscript{\rm 4}, Yongqi Tong\textsuperscript{\rm 5} \\
\textsuperscript{\rm 1}College of Electronics and Information Engineering, Sichuan University, China \\
\textsuperscript{\rm 2}School of Computer Science, The University of Adelaide, Australia \\
\textsuperscript{\rm 3}University of Electronic Science and Technology of China, China\  \  \textsuperscript{\rm 4}Tencent \\
\textsuperscript{\rm 5}College of Computer Science, Sichuan University, China\\
yanjie.gou@outlook.com, yinjie@scu.edu.cn, lingqiao.liu@adelaide.edu.au\\ daiyongya@yahoo.com, lineshen@tencent.com}

\begin{document}
\maketitle

\begin{abstract}
Aspect Sentiment Triplet Extraction (ASTE) aims to extract the spans of aspect, opinion, and their sentiment relations as sentiment triplets. 
Existing works usually formulate the span detection as a $1D$ \textit{token} tagging problem, and model the sentiment recognition with a $2D$ tagging matrix of \textit{token pairs}. Moreover, by leveraging the token representation of Pretrained Language Encoders (PLEs) like BERT, they can achieve better performance. 
However, they simply leverage PLEs as feature extractors to build their modules but never have a deep look at \textit{what specific knowledge does PLEs contain.}
In this paper, we argue that instead of further designing modules to capture the inductive bias of ASTE, PLEs themselves contain ``enough'' features for $1D$ and $2D$ tagging:
(1) The token representation contains the contextualized meaning of token itself, so this level feature carries necessary information for $1D$ tagging. 
(2) The attention matrix of different PLE layers can further capture multi-level linguistic knowledge existing in token pairs, which benefits $2D$ tagging. 
(3) Furthermore, with simple transformations, these two features can also be easily converted to the $2D$ tagging matrix and $1D$ tagging sequence, respectively. That will further boost the tagging results.
By doing so, \textit{PLEs can be natural tagging frameworks} and achieve a new state of the art, which is verified by extensive experiments and deep analyses.
\end{abstract}

\section{Introduction}
Sentiment Analysis \cite{Liu_2012,10.1145/2436256.2436274} is an important Natural Language Understanding task (NLU) to identify the sentiment from review sentences, which has been widely studied in many fields, e.g., E-commerce \cite{7975207} and social media \cite{10.5555/2021109.2021114}.
Recently, Aspect-based Sentiment Analysis \cite{pontiki-etal-2014-semeval} tries to perform sentiment analysis at a fine-grained level, which comprises several subtasks, such as Aspect Term Extraction \cite{10.5555/3304222.3304353}, Aspect Opinion Extraction \cite{fan-etal-2019-target}, and Aspect Sentiment Classification \cite{ruder-etal-2016-hierarchical}. 
In order to provide a unified solution for these subtasks, Aspect Sentiment Triplet Extraction (ASTE) is proposed by \cite{Peng_Xu_Bing_Huang_Lu_Si_2020} to extract sentiment triplets from review sentences, which contain all of the aspect terms, corresponding opinion spans, and their sentiment relations. For instance, given a review ``\textit{The ambience was nice but the service wasn’t so great .}'', the triplets of [\textit{ambience, nice, positive}] and [\textit{service, wasn’t so great, negative}] should be extracted.

\begin{figure}[t]
\begin{center}
\includegraphics[width=\columnwidth]{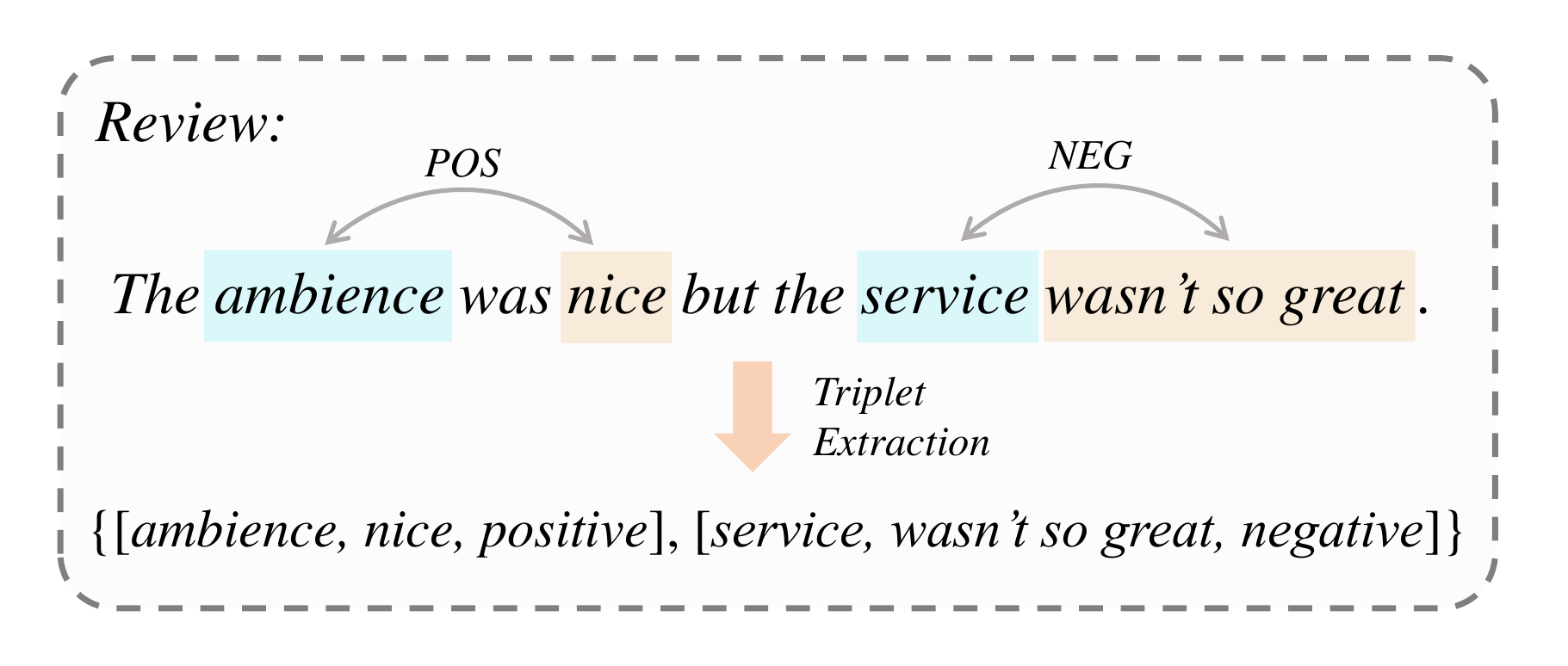}
\caption{An example of ASTE, which extracts the triplets (i.e., \textit{aspect terms, corresponding opinion spans, and their sentiment relations}) from the review sentence.
}
\label{fig:example}
\end{center}
\end{figure}

To recognize the triplet elements, many efforts are devoted. Most of the existing works design various modules to detect the spans of aspect and opinion, as well as the sentiment relations of them, which can be divided into two categories: (1) \cite{Peng_Xu_Bing_Huang_Lu_Si_2020,chen2021bidirectional} conduct ASTE in multiple stages, which firstly extract aspect terms and opinion spans, and then combine the valid pairs of them and decide their sentiment relations. 
(2) Some works \cite{xu-etal-2020-position,wu-etal-2020-grid,jing2021seeking} formulate ASTE in an end-to-end manner, by designing tagging schemes (i.e., 1$D$ token level tagging scheme \cite{xu-etal-2020-position} and 2$D$ token pair tagging scheme \cite{wu-etal-2020-grid,jing2021seeking}) to jointly extract the triplet elements. Furthermore, the aforementioned methods demonstrate that the adoption of Pretrained Language Encoders like BERT \cite{devlin-etal-2019-bert}, is beneficial for improvement.

\begin{figure}[t]
\begin{center}
\includegraphics[width=\columnwidth]{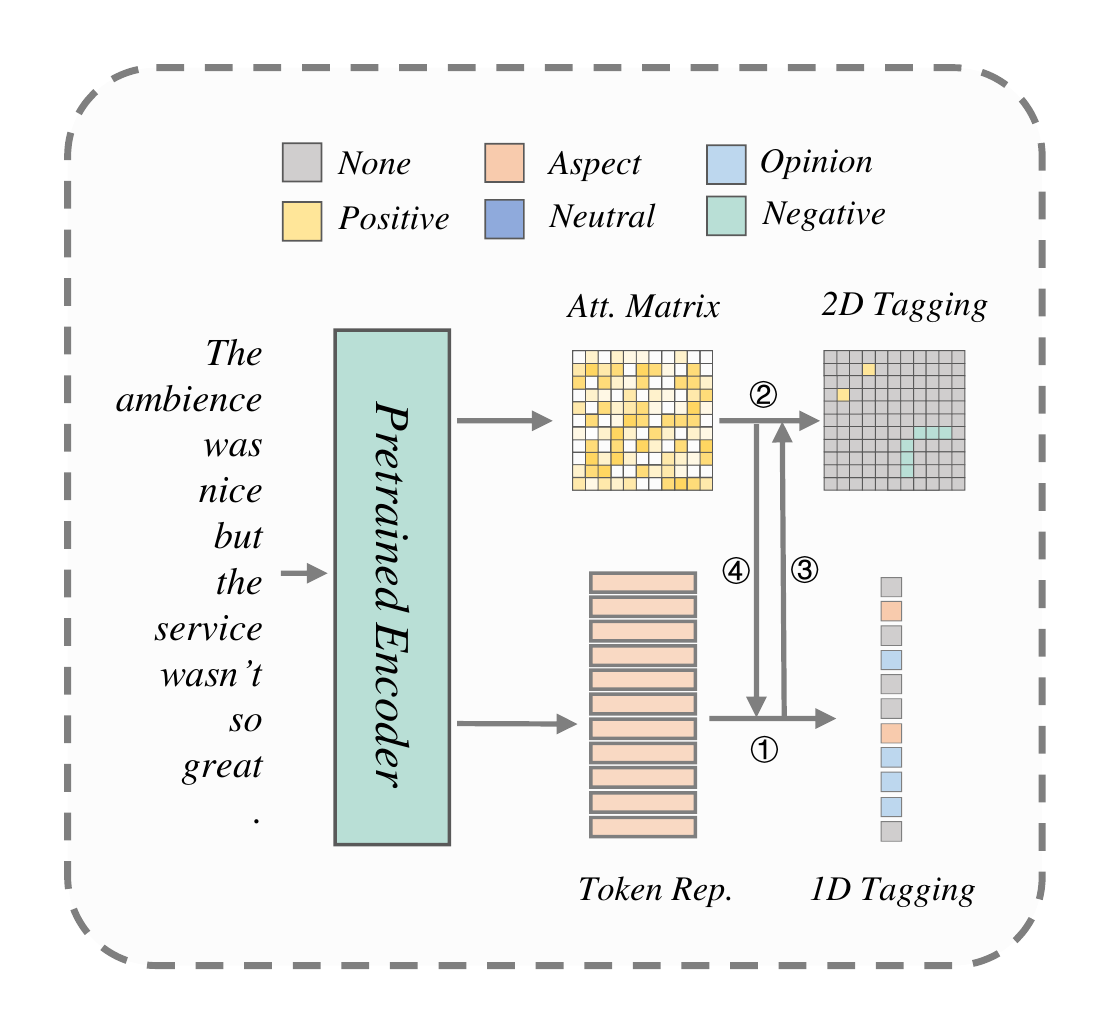}
\caption{An illustration of the natural tagging framework derived from PLEs, where $1D$ and $2D$ tagging schemes are converted from both the token representation and attention matrix (i.e., the branches \ding{172}-\ding{175}). In contrast, the existing works only leverage the branches \ding{172} and \ding{174} at most, \textit{which limits the potential of PLEs.}}
\label{fig:tagging}
\end{center}
\end{figure}

Specifically, they simply use the token representation of PLEs as a backbone of their designed modules to capture the inductive bias of ASTE, such as the span information of aspect and opinion \cite{xu-etal-2021-learning}. However, we argue that is not the optimal way to leverage PLEs for ASTE, since the knowledge stored in them, i.e., token representation and attention matrix, is not fully used. As shown in Fig. \ref{fig:tagging},
(1) The token representation contains the contextualized meaning of token itself, so this level feature carries necessary information to recognize aspect and opinion spans as a $1D$ tagging sequence (i.e., branch \ding{172} in Fig. \ref{fig:tagging}). 
(2) The attention matrix of different layers in PLEs can capture multi-level linguistic knowledge existing in the token pairs. As \cite{jawahar-etal-2019-bert} analyzed, the bottom layers focus more on phrase level syntactic information, and the top layers mainly capture semantic features. That means it contains effective features to recognize the sentiment relations between aspect and opinion spans with a $2D$ tagging matrix (i.e., branch \ding{173}).
(3) Besides, the token representation and attention matrix can also be converted to $2D$ tagging matrix and $1D$ tagging sequence by some simple transformations (i.e., branches \ding{174} and \ding{175}), so as to further boost the tagging results.

After these observations, we argue that \textit{PLEs themselves are naturally advanced tagging frameworks, due to the rich knowledge contained in both the token representation and \textbf{attention matrix}. All we need to do is converting them into the final tagging results.}
To this end, we propose \textbf{\textit{SimpleTag}} to fully leverage the knowledge stored in both the feature sources of PLEs.
Specifically, for the token representation, we use one branch to label the aspect and opinion spans with the token level tags of $\{A, O, N\}$, where $\{A, O\}$ means the token belongs to an \underline{A}spect or \underline{O}pinion span; $N$ means the token is \underline{N}ot one of the tokens of an aspect or opinion. 
Also, to detect the sentiment relations between the aspect and opinion, we leverage another branch to interact different token pairs in a multi-head selection manner \cite{bekoulis2018joint} with the token pair level tags $\{Pos, Neu, Neg, N\}$. $\{Pos, Neu, Neg\}$ means the token pair contains a \underline{Pos}itive, \underline{Neu}tral or \underline{Neg}ative sentiment relation if this token pair is from an aspect and opinion span respectively; $N$ means the token pair does \underline{N}ot contain a sentiment relation. \footnote{Our method is tagging-scheme-agnostic, which means more advanced $1D$ and $2D$ tagging schemes \cite{xu-etal-2020-position,wu-etal-2020-grid,jing2021seeking} can be easily plugged into our method for more complex tagging problem like nested ones with parts or all of the branches in Fig. \ref{fig:tagging}. Here we use a simple tagging scheme to highlight our motivation.}

For the attention matrix, we also use two branches to model the attention matrix, which can further boost the aspect and opinion tagging, as well as the sentiment relation recognition between them. That is, one branch labels the token pairs of the same words (i.e., the diagonal of attention matrix) with the tags $\{A, O, N\}$. Another branch assigns the token pair level features with the sentiment relation tags, i.e., $\{Pos, Neu, Neg, N\}$.
When the prediction is done, we apply a simple late fusion strategy to fuse the prediction logits to benefit our framework from both the knowledge of token representation and attention matrix. Finally, we use the fused result to decode sentiment triplets.

After conducting extensive experiments on four benchmarks \cite{xu-etal-2020-position}, we demonstrate PLEs themselves naturally contain ``enough'' token and token pair level knowledge for ASTE: By leveraging these features with only simple classification heads, 
SimpleTag can outperform the previous works and achieve a new state of the art.

To summarize, our contributions are as follows:
\begin{itemize}
    \item We explicitly leverage the attention matrix derived from PLEs to access the token pair level knowledge for ASTE.
    \item We propose SimpleTag, which is a natural tagging framework derived from PLEs themselves. By leveraging both the token and token pair level features, the rich knowledge can be fully mined to enhance the tagging results. 
    \item The experimental results on four public benchmarks demonstrate that our method can achieve a new state of the art.
\end{itemize}

\section{Related Works}

Aspect Sentiment Triplet Extraction is proposed by \cite{Peng_Xu_Bing_Huang_Lu_Si_2020}, which aims to extract the triplets of all the aspect terms, opinion spans and the sentiment relations between them.
To achieve that goal, many efforts are devoted.
\cite{Peng_Xu_Bing_Huang_Lu_Si_2020} proposes to extract the aspects and opinions at first, which will be combined into sentiment triplets later. 
\cite{chen2021bidirectional, mao2021joint} transform ASTE task into a Machine Reading Comprehension (MRC) task to capture the connections among the subtasks of ASTE.
\cite{huang2021target} proposes a two-stage method to enhance the correlations between aspects and opinions. 
\cite{jian2021aspect} proposes to regard the aspect and opinion terms as arguments of the expressed sentiment in a hierarchical reinforcement learning framework.
\cite{xu-etal-2021-learning} uses a span level approach to explicitly consider the interactions between the whole spans of aspects and opinions when predicting their sentiment relations.
Besides, \cite{xu-etal-2020-position,wu-etal-2020-grid,chen-etal-2021-semantic,jing2021seeking} propose unified tagging schemes to extract sentiment triplets in one stage:
\cite{xu-etal-2020-position} uses a token level tagging scheme, i.e., Position-aware Tagging Scheme, to extraction the sentiment triplets; 
\cite{wu-etal-2020-grid,chen-etal-2021-semantic} use a token pair level tagging scheme, which results in a 2$D$ tagging matrix. 
In addition, \cite{zhang-etal-2021-towards-generative,yan-etal-2021-unified} both propose to extract the sentiment triplets via a generative way, where a sequence-to-sequence paradigm is used.

\section{PLEs as Natural Tagging Frameworks}
In this Section, we first describe the overall workflow of SimpleTag.  Then, we elaborate on each component, i.e., Sentence Encoder, Tagging Layer (i.e., the branches \ding{172}-\ding{175} in Fig. \ref{fig:tagging}), and Triplet Decoding Procedure.

\subsection{Overall Workflow of SimpleTag} 
As show in Fig. \ref{fig:tagging}, after the Pretrained Language Encoder (e.g., BERT) derives the token representation $H$ and attention matrix $A$. We fully and explicitly leverage ALL the knowledge of them in the Tagging Layer. (1) Branch \ding{172} and \ding{175}: the aspect and opinion span are predicted with the 1$D$ token level tags $\{A, O, N\}$ by $T^{1D}_1 = \phi_1(H)$ and $T^{1D}_2 = \phi_2(A)$; (2) Branch \ding{173} and \ding{174}: the sentiment relation is predicted with the 2$D$ token pair level tags $\{Pos, Neu, Neg, N\}$ by $T^{2D}_1 = \theta_1(H)$ and $T^{2D}_2 = \theta_2(A)$. Then, we use a simple late fusion strategy to fuse the predicted logits of $T^{1D}$ and $T^{2D}$, so as to take advantage of both the token and token pair level knowledge of PLEs. Finally, the fused results are used to decode sentiment triplets.

\subsection{Sentence Encoder}
\label{subsec:Encoder}

Our Sentence Encoder aims to fully access the knowledge in both the token representation and attention matrix of PLEs, where these rich linguistic features of token level and token pair level can be fully mined. \textit{Here we choose BERT as the representative of them, in order to keep consistent with most of the previous works.}

Specifically, given one review sentence $\mathcal{S} = [w_1, w_2, ..., w_n]$, we first obtain its input embedding sequence. That is, $H_0 = [e_1, e_2, ..., e_n]$ ($e_i = w_i + p_i$), where $w_i$ and $p_i$ are the word embedding and position embedding of the $i$-th word. Then, the input embedding sequence is feed into BERT to obtain its token level representation and token pair level attention matrix:
\begin{align}
    H_i&, A_i = BERT\_Layer_i(H_{i-1}),  \\
    A^{1-L} = &[A_1; A_2; ...; A_{L-1}; A_L],  i \in [1, L]
\end{align}
where $A_i \in \mathbb{R}^{h \times n \times n}$ is the derived $h$ head attention matrix of $i$-th layers, and $A^{1-L} \in \mathbb{R}^{(L*h) \times n \times n}$) is the stacked attention matrix of all BERT layers.

Afterwards, the vanilla attention matrix may contain redundant and useless features \cite{michel2019sixteen}, which indicates we need to distill out the task-specific knowledge from it. To achieve this, we leverage several convolution blocks to model this 2$D$ matrix, which is a general way to refine the token pair level information and it has been applied in several NLP tasks like Incomplete Utterance Rewriting \cite{qian2020incomplete} and Document-level Relation Extraction \cite{ijcai2021-551}. 

Specifically, with the definition of one convolution block as:
\begin{align}
    X_{i-1}' = \sigma(Conv(X_{i-1})), \\
    X_i = BatchNorm(X_{i-1}'), 
\end{align}
where we conduct the convolution operation with the kernel size of $3\times3$ and use ReLU as the activation function $\sigma(\cdot)$. The channels of output are the same as the input. The refined process is as follows:
\begin{align}
    X_0 &= [A^{1-L}; R], i \in [1, C], \\
    X_i &= Conv\_Block(X_{i-1}),
\end{align}
where $R \in \mathbb{R}^{d_p \times n \times n}$ is the learnable parameters of relative position embeddings between token pairs. We use the final output $X_C$ as our refined attention matrix $A$, which is along with the token representation $H=H_L \in \mathbb{R}^{n \times d}$ as the provided features for the following Tagging Layer.

We argue that this is a more effective way to leverage the pretrained knowledge in PLEs, since the attention matrix of different layers originally store the rich linguistic knowledge via the pretraining paradigm \cite{jawahar-etal-2019-bert,clark-etal-2019-bert,DBLP:journals/corr/abs-1901-05287}. In contrast, all the existing works of ASTE only use the token representations of the last layer, which can result in losing task-specific features.

\subsection{Tagging Layer}
After the token and token pair level features are obtained, i.e., $H$ and $A$, they are leveraged to predict the aspect and opinion spans as a $1D$ tagging sequence and the sentiment relations between them as a $2D$ tagging matrix.

\subsubsection{Aspect and Opinion Recognition}
For aspect and opinion spans, we use the tags of $\{A, O, N\}$ to label the these features. 

Specifically, for $H$ and $A$, we implement $T^{1D}_1 = \phi_1(H)$ and $T^{1D}_2 = \phi_2(A)$ with two fully-connected layers to map them into the 1$D$ tag sequences $T^{1D}_1, T^{1D}_2 \in \mathbb{R}^{n \times 3}$, respectively. That is, 
\begin{align}
    T^{1D}_1 &= HW^{1D}_1 + b^{1D}_1, \\
    A' &= diagonal(A), \\
    T^{1D}_2 &= A'W^{1D}_2 + b^{1D}_2,    
\end{align}
where the $diagonal(\cdot)$ means taking the refined attention features between the same tokens. $W^{1D}_1, W^{1D}_2, b^{1D}_1$ and $b^{1D}_2$ are learnable parameters.

Finally, we use late fusion to sum both the prediction results, i.e., $T^{1D} = T^{1D}_1 + T^{1D}_2$ as the finally prediction to recognize aspect and opinion spans.

\subsubsection{Sentiment Relation Recognition} 
In the meanwhile, we also leverage $H$ and $A$ to recognize the sentiment relations between aspect and opinion by a 2$D$ token pair level tagging scheme $T^{2D} \in \mathbb{R}^{n \times n \times 4}$, where the classes belong to $\{Pos, Neu, Neg, N\}$.

Specifically, to convert the token representation $H$ to the $2D$ tagging matrix (i.e., $T^{2D}_1 = \theta_1(H)$), we implement it with the multi-head selection mechanism \cite{bekoulis2018joint}:
The $t$-th head is used to predict the $t$-th class between different token pairs. The detailed process for the $t$-th head is as follows:
\begin{align}
    &H^{q}_i = H_iW^{q}_t + b^{q}_t, t \in [1,4] \\
    &H^{k}_j = H_jW^{k}_t + b^{k}_t, \\
    &T^{2D,t}_{1,ij} = H^{q}_i R_{i-j} (H^{k}_j)^T,
\end{align}
where $T^{2D,t}_{1} \in \mathbb{R}^{n \times n}$ is the predicted tagging matrix for the $t$-th tag of $\{Pos, Neu, Neg, N\}$. Here we denote $T^{2D}_1 \in \mathbb{R}^{n \times n \times 4}$ as the whole-class prediction from the token representation $H$. $R_{i-j}$ means the relative position embedding between the $i$-th and $j$-th token, where we implement it with the rotary position embedding \cite{su2021roformer}.

In addition, to convert the refined attention matrix to the $2D$ tagging matrix (i.e., $T^{2D}_2 = \theta_2(A)$), we also implement this process with a fully-connected layer, which is as follows:
\begin{align}
    T^{2D}_2 = AW^{2D}_2 + b^{2D}_2,
\end{align}

Finally, we sum both prediction results to enhance the performance for sentiment relation recognition, i.e., $T^{2D} = T^{2D}_1 + T^{2D}_2$.

When training SimpleTag, we use Cross Entropy loss (CE) to supervise our model, which is as follows:
\begin{align}
    Loss = \sum_{k=1}^{n}CE(T^{1D}_{k}, y_k^{1D}) + \nonumber\\  \sum_{i=1}^{n}\sum_{j=1}^{n}CE(T_{i,j}^{2D}, y_{i,j}^{2D}),
\end{align}
where $T^{1D}_{k}$ and $T_{i,j}^{2D}$ denote the $k$-th predicted $1D$ tag and the predicted $2D$ tag between the $i$-th and $j$-th tokens. $y_k^{1D}$ and $y_{i,j}^{2D}$ denote the ground truth for $1D$ and $2D$ tagging.

\subsection{Triplet Decoding Procedure}
After the prediction and fusion, the result is used to decode the final sentiment triplets. (1) We first recognize the aspect and opinion span by searching the 1$D$ predicted tags, where the continuous tags of $A$ are recognized as an aspect and $O$ as an opinion span. (2) Then, we count the 2$D$ tags $\{Pos,Neu,Neg\}$ of the corresponding word pairs between the recognized aspect terms and opinion spans, where the most predicted tag is assigned as the sentiment relation of this triplet. \textit{Due to the symmetry of token pairs in the tagging matrix, we only use its upper triangle part for decoding.}

\section{Experiment}

\subsection{Datasets}
There are two versions of datasets for ASTE: ASTE-Data-V1 is released by  \cite{Peng_Xu_Bing_Huang_Lu_Si_2020} and ASTE-Data-V2 is released by \cite{xu-etal-2020-position}. They both include three datasets in the restaurant domain and one dataset in the laptop domain. 
However, ASTE-Data-V1 does not contain cases where one opinion span is associated with multiple targets, but these cases are very common in the real world. 
V2 refines the V1 version with these additional missing triplets and removes triplets with conflicting sentiments. Therefore, we use ASTE-Data-V2 for our experiments, which is in a more general setting. 

\subsection{Evaluation Metrics}
Following the existing works \cite{Peng_Xu_Bing_Huang_Lu_Si_2020,xu-etal-2020-position,wu-etal-2020-grid,chen2021bidirectional}, we use precision, recall, and F1 score as the metrics to evaluate the performance of ASTE. A correct triplet requires an exact match between the prediction of the aspect term, opinion span, and the sentiment polarity with the ground truth. 
Note that the F1 score takes into account both precision and recall, which can be regarded as a harmonic average of them. Therefore, we focus on the F1 score in the following experiments.

\subsection{Implementation Details}
The hyper-parameters in our experiment are tuned over the development set by grid search. We use \textit{bert-base-uncased} as our Sentence Encoder to be consistent with most of the previous works. The learning rate of all the parameters is set to $5e-5$ with gradient clip of 1.0, where the Adam optimizer \cite{DBLP:journals/corr/KingmaB14} is used for model optimization with a batch size of 16. Besides, the number of convolutional layers is selected from [2, 4, 6]. The dimension of the learnable relative position embeddings is set to 64.
Our implementation is based on PyTorch \cite{NEURIPS2019_bdbca288} and HuggingFace’s transformers library \cite{wolf-etal-2020-transformers}.

\begin{table*}[t]
\centering
\resizebox{2.0\columnwidth}{24mm}{%
\begin{tabular}{l|ccc|ccc|ccc|ccc}
\hline 
\multirow{2}{*}{Model} & \multicolumn{3}{c|}{Res14} & \multicolumn{3}{c|}{Lap14} & \multicolumn{3}{c|}{Res15} & \multicolumn{3}{c}{Res16} \tabularnewline
 & $P.$ & $R.$ & $F1$ & $P.$ & $R.$ & $F1$ & $P.$ & $R.$ & $F1$ & $P.$ & $R.$ & $F1$ \tabularnewline
\hline 
JET$^o_{\text{+}\text{BERT}}$ & 70.56 & 55.94 & 62.40 & 55.39 & 47.33 & 51.04 & 64.45 & 51.96 & 57.53 & 70.42 & 58.37 & 63.83  \tabularnewline
GTS$_{\text{+}\text{BERT}}$ & 67.76 & 67.29 & 67.50 & 57.82 & 51.32 & 54.36 & 62.59 & 57.94 & 60.15 & 66.08 & 69.91 & 67.93  \tabularnewline
(Jing et al., 2021)$_{\text{+}\text{BERT}}$ & 67.95 & 71.23 & 69.55 & 62.12 & 56.38 & 59.11 & 58.55 & 60.00 & 59.27 & 70.65 & 70.23 & 70.44  \tabularnewline
(Yan et al., 2021)$_{\text{+}\text{BART}}$ & 65.52 & 64.99 & 65.25 & 61.41 & 56.19 & 58.69 & 59.14 & 59.38 & 59.26 & 66.60 & 68.68 & 67.62  \tabularnewline
{Dual-MRC}$_{\text{+}\text{BERT}}^{\dag}$  & 71.10 & 70.11 & 70.60 & 58.52 & 54.86 & 56.63 & 64.84 & 54.06 & 58.96 & 67.40 & 68.37 & 67.88 \tabularnewline
{BMRC}$_{\text{+}\text{BERT}}^{\dag}$ & 70.12 & 70.40 & 70.26 & 66.24 & 53.64 & 59.28 & 62.20 & 59.56 & 60.85 & 68.42 & 70.32 & 69.36 \tabularnewline
UIE$_{\text{+}\text{T5}}$ & - & - & 71.27 & - & - & 58.69 & - & - & 59.60 & - & - & 70.24 \tabularnewline
Span-ASTE$_{\text{+}\text{BERT}}$ & 72.89 & 70.89 & \underline{71.85} & 63.44 & 55.84 & \underline{59.38} & 62.18 & 64.45 & \underline{63.27} & 69.45 & 71.17 & \underline{70.26} \tabularnewline
\hline
SimpleTag$_{\text{+}\text{BERT}}$ & 74.74 & 72.86 & \textbf{73.79} & 64.75 & 58.41 & \textbf{61.42} & 63.71 & 65.15 & \textbf{64.42} & 70.34 & 73.49 & \textbf{71.88} \tabularnewline
\hline 
\end{tabular}}
\caption{\label{tab:main_res}The overall evaluation results. $P.$ and $R.$ are Precision and Recall respectively. The best results are in \textbf{bold font} and the second-best ones are \underline{underlined}. $\dag$ indicates since \cite{mao2021joint,chen2021bidirectional} conduct experiments on ASTE-Data-V1, their results are reproduced on ASTE-Data-V2. ``-'' means the corresponding metric value is not reported in their original papers. Besides, to make fair comparison, the retrieved results of UIE are based on \textit{T5-v1.1-base} and without external data post-training.}
\end{table*}

\subsection{Compared Methods}
Our method is compare with the mostly recent states of the arts, which are as follows:
\begin{itemize}
    \item JET: \cite{xu-etal-2020-position} proposes to extract sentiment triplets by a position-aware tagging scheme.
    \item GTS: \cite{wu-etal-2020-grid} uses a grid tagging scheme to extract sentiment triplets.
    \item \cite{jing2021seeking}: this work proposes a jointly optimized dual-encoder model for ABSA to boost the performance of ABSA tasks.
    \item Dual-MRC: \cite{mao2021joint} proposes a dual-MRC framework to handle ASTE task, by jointly training two BERT-MRC models with parameter sharing.
    \item BMRC: \cite{chen2021bidirectional} proposes a bidirectional MRC framework to capture and utilize the associations among ASTE subtasks.
    \item \cite{yan-etal-2021-unified}: this work proposes a generative framework for ABSA.
    \item Span-ASTE: \cite{xu-etal-2021-learning} explicitly considers the interaction between the whole span of the aspect and opinion when predicting their sentiment.
    \item UIE: \cite{lu-etal-2022-uie} unifies Information Extraction (including ASTE) with the proposed structural schema instructor and structural extraction language.
\end{itemize}
\textit{What these methods have in common is that all of they only use the token representation to capture the features that ASTE needs, which ignores that the PLEs can become natural tagging frameworks by fully leveraging the token representation and \textbf{attention matrix}.}

\subsection{Overall Evaluation}

As reported in Tab. \ref{tab:main_res}, although these methods use Pretrained Models like BERT \cite{devlin-etal-2019-bert}, BART \cite{lewis2020bart} and T5 \cite{2020t5} in different ways to leverage its capability of deep language understanding and generation, our method can consistently outperform all of them. 
Specifically, compared to the recent state-of-the-art method Span-ASTE \cite{xu-etal-2021-learning}, although it use the token representation to create span level features, we can still outperform it by 1.94\%, 2.04\%, 1.15\% and 1.62\% on the four datasets respectively. 
In addition, although \cite{chen2021bidirectional,mao2021joint,yan-etal-2021-unified,lu-etal-2022-uie} use Pretrained Models in a MRC or generative way, so as to capture the discriminative features for ASTE, our method can also improve the performance by 2.52\%, 2.24\%, 3.57\% and 1.64\% on the four datasets, respectively.

The results demonstrate that our method SimpleTag, which treats PLEs as natural tagging frameworks, can makes full and explicit use of the pretrained knowledge in PLEs and is more effective to tackle ASTE and can achieve a new state of the art.
\section{Analysis and Discussion}
In this part, we make deep analyses of SimpleTag from both qualitative and quantitative perspectives. 

\subsection{Ablation Study}
We firstly conduct various ablated experiments to analyze the contributions of different features in SimpleTag, where each branch (\ding{172}-\ding{175}) in Fig. \ref{fig:tagging} is removed as a variant.

As reported in Tab. \ref{tab:ablation}, 
(1) When the branches from the attention matrix to $1D$ and $2D$ tagging are prohibited (i.e., \textit{w/o att. matrix}), our framework can only get features from the token representation. That makes it degrade to a variant which is similar to \cite{wu-etal-2020-grid}, where their performance is at a same level: the performance of this variant averagely drops by 4.34 points. \textit{That demonstrates that the attention matrix of PLEs does contain much richer task-specific features for ASTE, which can provide more effective information than the modules proposed by the existing works.}
(2) If one of the branches rooted from the attention matrix is removed (i.e., \textit{w/o att. matrix $\rightarrow$ $1D$ / $2D$ tag}), the performance of SimpleTag also declines by 2.15 and 3.27 points. That verifies the attention feature can boost both the aspect and opinion detection as well as sentiment relation recognition.

\begin{table*}[htb]
\centering
\resizebox{2.0\columnwidth}{25mm}{%
\begin{tabular}{l|ccc|ccc|ccc|ccc|c}
\hline 
\multirow{2}{*}{Model} & \multicolumn{3}{c|}{Res14} & \multicolumn{3}{c|}{Lap14} & \multicolumn{3}{c|}{Res15} & \multicolumn{3}{c|}{Res16} & \multicolumn{1}{c}{Avg}\tabularnewline
 & $P.$ & $R.$ & $F1$ & $P.$ & $R.$ & $F1$ & $P.$ & $R.$ & $F1$ & $P.$ & $R.$ & $F1$ & $\Delta F1$\tabularnewline
\hline
SimpleTag$_{\text{+}\text{BERT}}$ & 74.74 & 72.86 & \textbf{73.79} & 64.75 & 58.41 & \textbf{61.42} & 63.71 & 65.15 & \textbf{64.42} & 70.34 & 73.49 & \textbf{71.88} & - \tabularnewline
\hline
\textit{w/o att. matrix} & 69.92 & 70.84 & 70.38 & 61.59 & 53.05 & 57.00 & 56.46 & 61.24 & 58.75 & 64.84 & 71.54 & 68.03 & -4.34 \tabularnewline
\textit{w/o att. matrix $\rightarrow$ 1$D$ tag} & 70.55 & 71.54 & 71.04 & 60.74 & 58.04 & 59.36 & 62.19 & 62.06 & 62.13 & 71.03 & 69.79 & 70.40 & -2.15 \tabularnewline
\textit{w/o att. matrix  $\rightarrow$ 2$D$ tag} & 70.33 & 71.04 & 70.68 & 59.57 & 56.93 & 58.22 & 60.41 & 60.41 & 60.41 & 66.67 & 71.73 & 69.11 & -3.27 \tabularnewline
\hline
\textit{w/o token rep.} & 67.89 & 70.84 & 69.33 & 56.23 & 55.08 & 55.65 & 61.41 & 59.38 & 60.38 & 67.51 & 67.25 & 67.38 & -4.69 \tabularnewline
\textit{w/o token rep. $\rightarrow$ 1$D$ tag} & 72.11 & 71.24 & 71.68 & 56.91 & 57.12 & 57.01 & 62.02 & 63.30 & 62.65 & 69.75 & 71.93 & 70.83 & -2.34 \tabularnewline
\textit{w/o token rep. $\rightarrow$ 2$D$ tag} & 70.55 & 70.33 & 70.44 & 57.80 & 55.45 & 56.60 & 62.45 & 60.00 & 61.20 & 70.30 & 67.84 & 69.05 & -3.56 \tabularnewline
\hline
\textit{w/o convolution} & 72.05 & 72.05 & 72.05 & 59.54 & 57.67 & 58.59 & 62.61 & 60.41 & 61.49 & 70.44 & 71.54 & 70.99 & -2.10 \tabularnewline
\textit{w/o rel. pos. emb.} & 70.46 & 73.16 & 71.78 & 58.95 & 62.11 & 60.49 & 60.62 & 62.32 & 61.44 & 71.23 & 69.01 & 70.10 & -1.93 \tabularnewline
\textit{w/o rotary pos. emb.} & 73.10 & 70.74 & 71.90 & 60.85 & 61.18 & 61.01 & 64.24 & 61.86 & 63.03 & 70.21 & 70.76 & 70.49 & -1.27 \tabularnewline
\hline 
\end{tabular}}
\caption{\label{tab:ablation}The ablation study of our method.}
\end{table*}

In addition, (3) When the two branches rooted from the token representation are all removed (i.e., \textit{w/o token rep.}), the performance drop by 4.69\%, which means the token representation also contains important information for ASTE. (4) When only one of these two branches is employed, the performance of SimpleTag can decline by 2.34\% and 3.56\%, which can draw a similar conclusion as (2).

Besides, (5) When we remove the convolution blocks (i.e., \textit{w/o convolution}), its performance drops by 2.1\%. That means the convolution operation benefits for the refinement of attention matrix, where the task-specific features can be effectively distilled out from them and fully boost ASTE. (6) We also remove the additional position embeddings so as to analyze their impacts on ASTE. As shown in the bottom of Fig. \ref{fig:tagging}, when the relative position embedding and rotary position embedding are removed (i.e., \textit{w/o rel. pos. emb.} and \textit{w/o rotary pos. emb.}), the F1 score also decline by 1.93\% and 1.27\%. That indicates both the position embeddings are useful for ASTE.

Therefore, the ablation study demonstrates the state-of-the-art performance can be achieved by simply leverage the features of PLEs with several classification heads to $1D$ and $2D$ tagging matrix.

\subsection{Effects of Attention in Different Layers}
To further investigate the effects of attention matrices in different layers of BERT, we also remove its low layers (1-4), middle layers (5-8) and high layers (9-12) during model training and inference. As shown in Tab. \ref{tab:layereffects}: When we remove the low, middle and high layers, the performance of SimpleTag averagely drops by 0.94\%, 1.71\% and 2.65\%, respectively. That demonstrates that the layers of different levels do contain task-specific features and all of them contribute to the improvement. That corresponds to \cite{jawahar-etal-2019-bert}, which reveals the the low, middle and high layers of PLEs contain useful knowledge, i.e., the phrase, syntactic and semantic level information, respectively. 

\begin{table}[htb]
\centering
\resizebox{0.95\columnwidth}{11mm}{%
\begin{tabular}{lccccc}
\hline 
Dataset & Res14 & Lap14 & Res15 & Res16 & $\Delta$\tabularnewline
\hline
\textit{w/ all layers} & \textbf{73.79} & \textbf{61.42} & \textbf{64.42} & \textbf{71.88} & -\tabularnewline
\hline
\textit{w/o 1-4 layers} & 72.69 & 59.94 & 64.19 & 70.92 & -0.94 \tabularnewline
\textit{w/o 5-8 layers} & 72.11 & 58.54 & 63.09 & 70.69 & -1.71 \tabularnewline
\textit{w/o 9-12 layers} & 71.58 & 58.43 & 61.60 & 69.30 & -2.65 \tabularnewline
\hline 
\end{tabular}}
\caption{\label{tab:layereffects}The effects (F1 scores) of attention matrices in different layers of BERT.}
\end{table}

\subsection{Effects of Different Pretrained Models}

\begin{table}[htb]
\centering
\resizebox{0.95\columnwidth}{8.8mm}{%
\begin{tabular}{lccccc}
\hline 
Dataset & Res14 & Lap14 & Res15 & Res16 & $\Delta$ \tabularnewline
\hline
BERT & 73.79 & 61.42 & 64.42 & \underline{71.88} & - \tabularnewline
XLNet & \underline{74.00} & \underline{62.08} & \textbf{67.20} & 71.29 & +0.77 \tabularnewline
ELECTRA\  \  \  \  \  \  \  \  \  & \textbf{75.13} & \textbf{62.58} & \underline{64.44} & \textbf{75.02} & +1.42 \tabularnewline
\hline 
\end{tabular}}
\caption{\label{tab:diffples}The effects of SimpleTag with different Pretrained Models. We use the \textit{base version} for all of them.}
\end{table}

Furthermore, instead of only applying our framework with the masked language model BERT, we also replace it with XLNet \cite{yang2019xlnet}, ELECTRA \cite{clark2020electra}, where the former is permutation language model and the latter is trained in a generator-discriminator way.
By doing so, we can verify the adaptability of our framework to other different pretraining paradigms.
As shown in Tab. \ref{tab:diffples}, compared with BERT-based Encoder, XLNET and ELECTRA can help our method achieve better performance with the improvements of 0.77\% and 1.42\%. That demonstrates our model exhibits good robustness to different pretraining paradigms.

\begin{figure*}[htb]
\begin{center}
\includegraphics[width=2.\columnwidth]{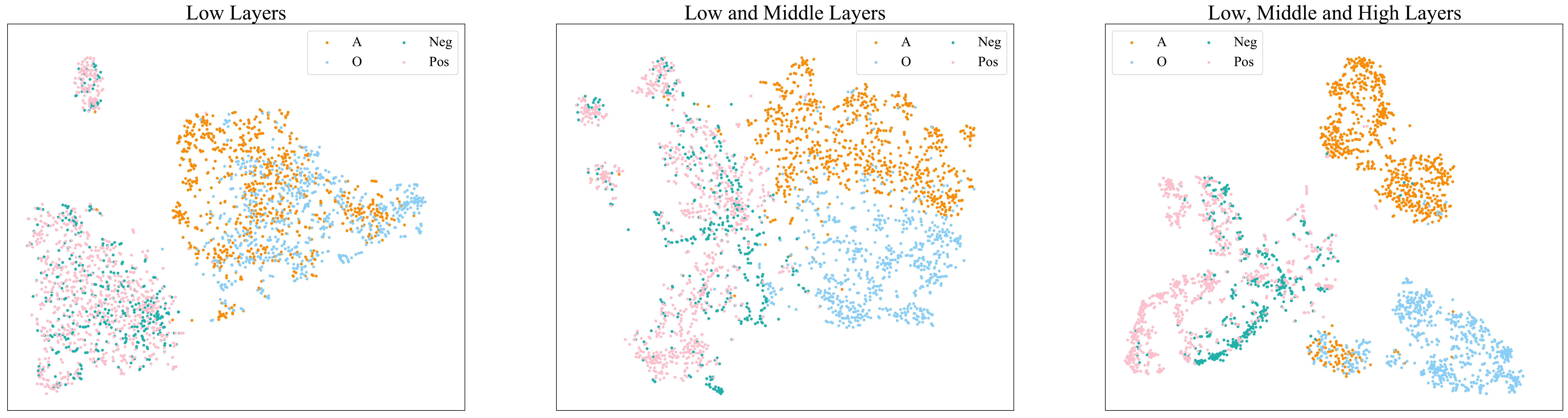}
\caption{The t-SNE visualization of the attention scores of different token pairs on Res15 dataset. We only visualize the representations whose classes belong to $\{$$A, O$$\}$ and $\{$$Pos, Neg$$\}$. The label $Neu$ is omitted since its number is too small in the training set (i.e., only 25 triplets) and can not be trained well, which can interfere with observation.
}
\label{fig:vis}
\end{center}
\end{figure*}

\subsection{Qualitative Visualization of Attention}

Besides, to demonstrate the knowledge stored in the attention matrices of different layers is beneficial to obtain informative and discriminative representations for ASTE, we also apply t-SNE \cite{JMLR:v9:vandermaaten08a} to these attention scores of different token pairs, and plot their 2-dimensional vectors. Since the residual connection in PLEs can potentially deliver the information of attention matrix from low layers to high layers, we choose to gradually concatenate the attention scores from low to high layers, where the difference between two visualization results can represent the ``real effect'' of corresponding layers, i.e., the low, middle and high layers.

It is obvious that (1) Only using the bottom layers can easily tell the difference between the classes of $\{A, O\}$ and $\{Pos, Neu, Neg\}$ with a large margin. That indicates the bottom layers do capture some task-specific information existing in sentiment triplets.
(2) By adding the attention scores of middle layers, we can observe that these features can further distinguish $O$ from $A$ with different clusters. 
(3) When the attention scores of all layers are used, it results in more compact clusters and clearer boundaries between the classes $O$ and $A$, and the sentiment classes of $Pos$ and $Neg$ can also be further recognized. 
That suggests the features in different layers all contribute to the performance, which are helpful to decide the classes the token pairs belong to. Without any part of them can result in the situation of losing task-specific information.

\subsection{Error Analysis}
To guide the future works with deep insights, we also conduct error analysis on Res14 to investigate what wrong decisions are made by SimpleTag.

The main incorrect triplets are divided into four categories. Most of the errors decoded by our method are {Span Detection Errors} (43.8\%). For example, give the sentence ``\textit{the french fries \textendash{} with the kaimata dip were terrific!}'', the aspect span comprised of multiple tokens (i.e., ``\textit{the french fries -- with kaimata dip}'') is wrongly predicted as two independent aspect terms as ``\textit{the french fries}'' and ``\textit{kaimata dip}''. This kind of mistake can be further solved by employing more effective convolution networks like U-Net \cite{RFB15a} to model the attention matrix, so as to detect span-level information \cite{xu-etal-2021-learning}.

Also, 35.6\% of the errors are introduced by {misclassifying the sentiment relations}. For the sentence ``\textit{dessert was also to die for!}'', our method predicts the sentiment as negative, which may be caused by the incorrect clue ``\textit{die}''. A solution for this mistake can be addressed by further adding phrase level information \cite{phrasebertwang2021}. 

In addition, in the wrongly decoded triplets, 15.2\% of them are actually correct but they are not annotated in the ground truths, which means {the datasets are not fully annotated}. For ``\textit{the sauce is excellent (very fresh) with dabs of real mozzarella.}'', one of our predicted triplets is ``(\textit{mozzarella, real, positive})'', which is correct but it is not annotated in the dataset.  

Beside, some sentences require more powerful capability of language understanding to distinguish which triplets should not be extracted. For example, given ``\textit{I came to fresh expecting a great meal, but all I got ...}'', our method wrongly predicts it with the triplet ``(\textit{meal, great, positive})''. For this kind of error, more effective method is required to capture these semantic level information.

\section{Limitations and Potential Risks}
In this paper, SimpleTag is only evaluated with several PLEs for ASTE, where the Pretrained Language Decoder like GPT-x \cite{Radford2018ImprovingLU,radford2019language,NEURIPS2020_1457c0d6} and T5 \cite{2020t5} are not explored and is left for future work. 
Besides, this paper demonstrates the effectiveness of SimpleTag on ASTE, whether other tagging-based task like Named Entity Recognition \cite{li2021unified} and Relation Extraction \cite{ijcai2021-551} can achieve the same performance improvement is at risk, which will be explored in the future.

\section{Conclusion}
In this work, we propose SimpleTag, a simple-yet-effective tagging framework which is naturally derived from PLEs themselves. By conducting various experiments, we demonstrate that, compared with the existing works which further design modules to capture the inductive bias of ASTE, the token representation and attention matrix of PLEs contain ``enough'' information for ASTE. Fully leveraging these features with only several simple transformations can further make SimpleTag outperform all the existing works and  obtain state-of-the-art performance on ASTE. 

\bibliography{anthology,custom}




\end{document}